\documentclass[10pt, twocolumn, letterpaper]{article}

\usepackage[accsupp]{axessibility} 

\usepackage{cvpr}              
\usepackage{graphicx}
\usepackage{amsmath}
\usepackage{amssymb}
\usepackage{booktabs}
\usepackage[numbers, sort&compress]{natbib}
\usepackage{multirow}

\usepackage[pagebackref, breaklinks, colorlinks]{hyperref}

\usepackage[capitalize]{cleveref}
\crefname{section}{Sec.}{Secs.}
\Crefname{section}{Section}{Sections}
\Crefname{table}{Table}{Tables}
\crefname{table}{Tab.}{Tabs.}

\def\cvprPaperID{*****} 
\def\confName{CVPR}
\def\confYear{2025}

\begin{document}

\title{Emotion Recognition with CLIP and Sequential Learning}

\author{Weiwei Zhou, Chenkun Ling, Zefeng Cai\\
China Telecom Cloud\\
{\tt \small \{zhouweiwei,lingchengk,caizf\}@chinatelecom.cn}
}

\maketitle

\begin{abstract}
Human emotion recognition plays a crucial role in facilitating seamless interactions between humans and computers. In this paper, we present our innovative methodology for tackling the Valence-Arousal (VA) Estimation Challenge, the Expression Recognition Challenge, and the Action Unit (AU) Detection Challenge, all within the framework of the 8th Workshop and Competition on Affective Behavior Analysis in-the-wild (ABAW).

Our approach introduces a novel framework aimed at enhancing continuous emotion recognition. This is achieved by fine-tuning the CLIP model with the aff-wild2 dataset, which provides annotated expression labels. The result is a fine-tuned model that serves as an efficient visual feature extractor, significantly improving its robustness. To further boost the performance of continuous emotion recognition, we incorporate Temporal Convolutional Network (TCN) modules alongside Transformer Encoder modules into our system architecture. The integration of these advanced components allows our model to outperform baseline performance, demonstrating its ability to recognize human emotions with greater accuracy and efficiency.
\end{abstract}
\section{Introduction}
\label{sec:intro}

Facial Expression Recognition (FER) holds immense potential across various domains, including emotion detection in videos, enhancing security through facial recognition, and enriching virtual reality experiences. While significant progress has been made in tasks such as face and attribute recognition, the challenge of accurately understanding emotions remains a complex problem.

Emotional expressions often involve subtle nuances, making it difficult to interpret emotions with precision and leading to ambiguity or uncertainty in assessment. This complexity poses challenges in accurately evaluating an individual’s emotional state. A key obstacle is the limited scope of current FER datasets, which fail to capture the full spectrum of human emotional expressions, hindering the development of effective models. To enhance the performance and reliability of FER systems, it is crucial to expand and diversify these datasets.

The Aff-Wild and Aff-Wild2 datasets, along with their corresponding challenges \cite{kollias20247th,kollias20246th,kollias2023abaw2, kollias2023multi, kollias2019expression, kollias2022abaw, kollias2021analysing, kollias2020analysing, kollias2021distribution, kollias2021affect, kollias2019face, kollias2019deep, zafeiriou2017aff, 2303.01498}, have played a pivotal role in advancing the field of affective recognition. The Aff-Wild2 dataset includes approximately 600 videos and 3 million frames, annotated with three key affective attributes: a) dimensional affect, including valence and arousal; b) six basic categorical affects; and c) facial action units. To encourage further utilization of the Aff-Wild2 dataset, the 8th ABAW\cite{Kollias2025} competition was organized, focusing on affective behavior analysis in natural, real-world environments.

Building upon the notable success of CLIP, we previously investigated its application as a fine-tuned visual feature extractor for facial expression datasets. Subsequently, we integrated Temporal Convolutional Networks (TCN) and Transformer models into our framework for continuous emotion recognition. This methodology yielded substantial improvements in evaluation accuracy across the tasks of Valence-Arousal Estimation, Action Unit Detection, and Expression Recognition.

The structure of the paper is organized as follows: Section \ref{sec:RelatedWork} provides a review of the literature on facial emotion recognition. Section \ref{sec:method} outlines the proposed methodology. Section \ref{sec:experiment} details the experimental setup and presents the results. Finally, Section \ref{sec:conclusion} offers the conclusions drawn from the study.

\section{Related Work}
\label{sec:RelatedWork}

Prior research has introduced several effective network architectures for the Aff-Wild2 dataset. Kuhnke et al. \cite{kuhnke2020two} integrated visual and audio information from videos to develop a two-stream network for emotion recognition, demonstrating superior performance. Similarly, Yue Jin et al. \cite{jin2021multi} proposed a transformer-based model to fuse audio and visual features for enhanced recognition accuracy.

NetEase \cite{zhang2023multi} leveraged visual information extracted from a Masked Autoencoder (MAE) model, pre-trained in a self-supervised manner on a large-scale face image dataset. Subsequently, the MAE encoder was fine-tuned on the image frames from the Aff-Wild2 dataset for the tasks of AU, Expression Recognition, and VA Estimation, which can be characterized as static and unimodal training. Furthermore, multi-modal and temporal information from the video data was incorporated, and a transformer-based framework was employed to effectively fuse these multi-modal features. SituTech \cite{liu2023multi} employed multi-modal feature combinations, derived from various pre-trained models, to capture more comprehensive and effective emotional information.

Temporal Convolutional Networks (TCN) were introduced by Colin Lea et al. \cite{lea2016temporal}, which effectively capture relationships across multiple time scales, including low-, intermediate-, and high-level temporal dependencies. Building on this, Jin Fan et al. \cite{fan2021parallel} proposed a model incorporating a spatial-temporal attention mechanism, designed to capture dynamic internal correlations, with stacked TCN backbones extracting features from various window sizes.

The Transformer mechanism, first introduced by Vaswani et al. \cite{vaswani2017attention}, has demonstrated exceptional performance across numerous tasks, prompting many researchers to apply it in the domain of affective behavior analysis. Zhao et al. \cite{zhao2021former} developed a model utilizing spatial and temporal Transformers for facial expression analysis. Additionally, Jacob et al. \cite{9577264} proposed a network employing a transformer correlation module to learn the relationships between action units.

Building upon these previous works, this paper proposes the use of CLIP as a feature extractor and introduces a model comprising TCN and Transformer components to enhance emotion recognition performance.

\section{Methodology}
\label{sec:method}

In this section, we provide a comprehensive description of our proposed method for addressing the three challenging tasks of affective behavior analysis in the wild, as outlined in the 8th ABAW Competition: Valence-Arousal Estimation, Expression Recognition, and Action Unit Detection. We detail the design of our model architecture, the data processing techniques employed, and the training strategies implemented for each task.

\subsection{Fine-tuning CLIP}

New fully connected layers are incorporated into the CLIP encoder. During the fine-tuning process, all parameters are updated to optimize the alignment of the feature extractor with the distribution of the Aff-Wild2 dataset. This adjustment ensures that the learned features of the model are better suited to the specific characteristics of the data, thereby enhancing its performance in emotion recognition tasks.

\subsection{Temporal Convolutional Network}

The videos are initially divided into segments with a window size $w$ and stride $s$. Given the segment window $w$ and stride $s$, a video with $n$ frames would be split into $[n/s] + 1$ segments, where the $i$-th segment contains frames$\left\{F_{(i-1) *s+1}, \ldots, F_{(i-1) * s+w}\right\}$.

In other words, the videos are divided into overlapping chunks, each containing a fixed number of frames. This approach serves to decompose the video into smaller, more manageable segments that are easier to process and analyze. Each chunk overlaps with the previous and subsequent ones to ensure that no information in the video is lost, thereby maintaining continuity across the segments.

We denote visual features as  $f_i$ corresponding to the $i$-th segment extracted by pre-trained and fine-tuned ViT-Base encoder.

The visual feature is fed into a dedicated Temporal Convolutional Network (TCN) for temporal encoding, which can be formulated as follows:
$$ g_i=\text { TCN }\left(f_i\right) $$

This implies that we utilize a specialized type of neural network capable of capturing the temporal patterns and dependencies of the features over time. The Temporal Convolutional Network (TCN) processes the input feature vector by applying a series of convolutional layers, each with varying kernel sizes and dilation rates, to generate an output feature vector. The output vector maintains the same length as the input but encodes richer information about the temporal context. For instance, the TCN can learn how the visual features evolve over time across each segment of the video, thus capturing the dynamic temporal changes inherent in the video data.

\subsection{Temporal Encoder}
We also employ a Transformer encoder to model the temporal information within the video segment, which can be formulated as follows:

$$ h_i=\text { TransformerEncoder }\left(g_i\right). $$ 

The Transformer encoder models the context within a single segment, thereby neglecting the dependencies between frames across different segments. To address this limitation and capture the inter-segment dependencies, overlapping between consecutive segments can be utilized. This ensures that the temporal relationships between frames across segments are preserved, which implies that $s \leq w$, where $s$ is the stride and $w$ is the window size.

We employ a different type of neural network capable of learning the relationships and interactions among features within each segment. The Transformer encoder processes the input feature vector by applying a series of self-attention layers followed by feed-forward layers, resulting in an output feature vector that holds more semantic meaning and representational power than the input. For instance, the Transformer encoder can learn the spatial relationships between different parts of the image within each segment of the video. However, the Transformer encoder operates within the confines of a single segment, neglecting the interdependencies between different segments of the video. To address this limitation, we introduce overlapping between consecutive segments, such that certain frames are shared by multiple segments. This approach allows the model to capture the temporal relationships and dependencies across segments, thereby enabling a more holistic understanding of the video’s content. The degree of overlap is determined by two parameters: $s$, the length of a segment, and $w$, the sliding window size. When $s$ is smaller than or equal to $w$, overlap between consecutive segments occurs, allowing the model to capture temporal dependencies across segments.

\subsubsection{Prediction}
After the temporal encoder, the features $h_i$ are passed through a Multi-Layer Perceptron (MLP) for regression, which can be formulated as follows:
$$ y_i= \text{MLP} (h_i) $$
where $y_i $ are the predictions of $i$-th segment. For VA challenge, $y_i \in \mathbb{R}^{l \times 2}$. For Expr challenge, $y_i \in \mathbb{R}^{l \times 8}$. For AU challenge, $y_i \in \mathbb{R}^{l \times 12}$ .

The prediction vector contains the estimated values for each segment. The Multi-Layer Perceptron (MLP) consists of several layers of neurons that learn non-linear transformations of the input features. The MLP is trained to minimize the error between the prediction vector and the ground truth vector. The ground truth vector represents the actual values we aim to predict for each segment. Depending on the specific challenge, the structure of the ground truth and prediction vectors varies.

For the Valence-Arousal (VA) challenge, we predict two values: valence and arousal. Valence indicates the degree of positivity or negativity of an emotion, while arousal reflects the level of activity or passivity of an emotion. In the Expression (Expr) recognition challenge, we predict eight values, each corresponding to one of the basic facial expressions: anger, disgust, fear, happiness, sadness, surprise, neutral, and other expressions. For the Action Unit (AU) detection challenge, we predict twelve values, one for each action unit (AU1, AU2, AU4, AU6, AU7, AU10, AU12, AU15, AU23, AU24, AU25, and AU26).

\begin{table*}
  \centering
  \begin{tabular}{@{}lcccccccc@{}}
    \toprule
    Task & Evaluation Metric & Partition & Method & Fold 0 & Fold 1 & Fold 2 & Fold 3 & Fold 4 \\
    \midrule
    \multirow{2}{*}{Valence} & \multirow{2}{*}{CCC} & \multirow{2}{*}{Validation} & Ours & 0.5615 & 0.6102 & 0.5025 & 0.5712 & 0.5504 \\
    & & & Baseline & 0.24 & - & - & - & - \\
    \cline{3-9}
    \multirow{2}{*}{Arousal} & \multirow{2}{*}{CCC} & \multirow{2}{*}{Validation} & Ours & 0.6125 & 0.5782 & 0.6231 & 0.6712 & 0.6298 \\
    & & & Baseline & 0.20 & - & - & - & - \\
    \midrule
    \multirow{2}{*}{Expr} & \multirow{2}{*}{F1-score} & \multirow{2}{*}{Validation} & Ours & 0.4651 & 0.4274 & 0.4562 & 0.4485 & 0.4702 \\
    & & & Baseline & 0.23 & - & - & - & - \\
    \midrule
    \multirow{2}{*}{AU} & \multirow{2}{*}{F1-score} & \multirow{2}{*}{Validation} & Ours & 0.5801 & 0.5461 & 0.5120 & 0.5642 & 0.5719 \\
    & & & Baseline & 0.39 & - & - & - & - \\
    \bottomrule
  \end{tabular}
  \caption{Results for the five folds of three tasks}
  \label{tab:results_fold}
\end{table*}

\subsection{Loss Functions}
For the VA challenge, we use the Concordance Correlation Coefficient (CCC) between the predictions and the ground truth labels as the evaluation metric. CCC measures the correlation between two sequences $x$ and $y$, and its value ranges between -1 and 1. A value of -1 indicates perfect anti-correlation, 0 represents no correlation, and 1 signifies perfect correlation. The loss function is then calculated as:

$$
CCC(x, y) = \frac{2 * \operatorname{cov}(x, y)}{\sigma_x^2+\sigma_y^2+\left(\mu_x-\mu_y\right)^2}
$$

\text { where } $\operatorname{cov}(x, y)=\sum\left(x-\mu_x\right) *\left(y-\mu_y\right)$

$$\mathcal{L}_{\text {VA }}=1-CCC $$

For the Expr challenge, we use the cross-entropy loss as the loss function. Cross-entropy loss is commonly used for classification tasks, where it measures the difference between the predicted probability distribution and the true distribution (ground truth):
$$
\mathcal{L}_{\text {EXPR }} = - \frac {1} {N}\sum_ {i} \sum_ {c=1}^My_ {ic}\log (p_ {ic})
$$

where $y_{ic}$ is a binary indicator (0 or 1) if class $c$ is the correct classification for observation $i$.
$p_{ic}$ is the predicted probability of observation $i$ being in class $c$, 
$M$ is the number of classes.
The multiclass cross-entropy loss function evaluates how effectively a model predicts the true probabilities for each class in a given observation. It penalizes incorrect predictions by computing the logarithm of the predicted probabilities. A lower loss indicates better model performance, as it signifies that the predicted probabilities are closer to the true probabilities for each class.

For the AU challenge, we employ BCEWithLogitsLoss as the loss function, which combines a sigmoid layer with binary cross-entropy.  BCEWithLogitsLoss is particularly effective for binary classification tasks, where each action unit is treated as an independent binary classification problem. It calculates the binary cross-entropy loss after applying the sigmoid activation function to the logits, allowing the model to output probabilities for each action unit:

$$
\mathcal{L}_{\text {AU }} = - \frac {1} {N}\sum_ {i} [y_i\cdot log (\sigma (x_i)) + (1-y_i)\cdot log (1-\sigma (x_i))]
$$

where $N$ is the number of samples, $y_i$ is the target label for sample $i$, $x_i$ is the input logits for sample $i$, $\sigma$ is the sigmoid function.

The advantage of using BCEWithLogitsLoss over BCELoss with a separate sigmoid function is that it combines the sigmoid activation and binary cross-entropy loss into a single, more stable operation. This integration helps to avoid numerical instability, particularly when dealing with extreme values in the logits, thereby improving both the model’s performance and training efficiency.

\section{Experiments and Results}
\label{sec:experiment}

\subsection{Experiments Settings}
All models were trained on two Nvidia GeForce GTX 3090 GPUs, each equipped with 24GB of memory.
\subsubsection{CLIP Fine Turning}

In the fine-tuning stage of CLIP, we adjusted the batch size to 256 and lowered the learning rate to 0.0001,  leveraging the AdamW optimizer.

\subsubsection{Task Training}

We employed the AdamW optimizer along with a cosine learning rate schedule, incorporating a warmup during the first epoch. The learning rate was set to $3e-5$, the weight decay to $1e-5$, the dropout probability to 0.3, and the batch size to 32.

For all three challenges, videos were split using a segment window of $w=300$ and a stride of $s=200$. This configuration divided each video into segments of 300 frames, with an overlap of 100 frames between consecutive segments. This strategy enabled the model to capture the temporal dynamics of facial expressions and emotions more effectively.

\subsection{Overall Results}

Table \ref{tab:results_fold} presents the experimental results of our proposed method on the validation set for the VA, Expr, and AU Challenges. The Concordance Correlation Coefficient (CCC) is used as the evaluation metric for valence and arousal prediction, while the F1-score is applied to evaluate the results of the Expr and AU challenges. As shown in the table, our proposed method significantly outperforms the baseline. These results highlight the effectiveness of our approach, which combines Temporal Convolutional Networks (TCN) and Transformer-based models to integrate visual and audio information, leading to improved accuracy in emotion recognition on this dataset.

\subsection{Ablation Studies}

\begin{table}
  \centering
  \begin{tabular}{@{}lccc@{}}
    \toprule
    Method & VA & Expr & AU \\
    \midrule
    baseline & 0.220 & 0.230 &0.390 \\
    w/o. Fine-tuning & 0.5049& 0.4123 &0.5274 \\
    w/o. TCN & 0.5698& 0.4473 &0.5621 \\
    w/o. Temporal Encoder & 0.5712& 0.4528 &0.5724 \\
    ours    & 0.5870 & 0.4651 & 0.5801 \\
    \bottomrule
   \end{tabular}
   \caption{ Ablation studies that discuss the significance of CLIP fine-tuning, Temporal Convolutional Network(TCN), and Temporal Encoder.}
   \label{tab:ablation}
\end{table}
In this section, we present a series of experiments to investigate the significance of each module in our approach, including CLIP fine-tuning, the Temporal Convolutional Network (TCN), and the Temporal Encoder. All experiments were conducted using the official training and validation sets. The results of these experiments are summarized in Table \ref{tab:ablation}.

\textbf{CLIP fine-tuning.} 
To assess the effectiveness of CLIP fine-tuning, we conducted an experiment where the fine-tuning step was removed, and we used the pre-trained CLIP encoder to directly extract features. The results show a significant decrease in the accuracy of all tasks. Specifically, the average CCC for VA dropped from 0.5870 to 0.5049, the average F1 score for Expr decreased from 0.4651 to 0.4123, and the average F1 score for AU reduced from 0.5801 to 0.5274. These findings indicate that CLIP fine-tuning plays a crucial role in leveraging static visual features from individual images, thus providing valuable prior knowledge that aids in learning temporal visual features for improved performance.

\textbf{Temporal Convolutional Network.} 
To investigate the importance of the Temporal Convolutional Network (TCN), we removed it from the model and observed a noticeable decline in the performance metrics across all tasks. The average CCC for VA decreased from 0.5870 to 0.5698, the average F1 score for Expr dropped from 0.4651 to 0.4473, and the average F1 score for AU fell from 0.5801 to 0.5621. These results demonstrate the effectiveness of the Temporal Convolutional Network in capturing temporal dependencies and highlighting its importance in sequence modeling.

\textbf{Temporal Encoder.} We also conducted an experiment by removing the Transformer Encoder from the model. The results showed a decline in performance across all tasks: the average CCC for VA decreased from 0.5870 to 0.5712, the average F1 score for Expr dropped from 0.4651 to 0.4528, and the average F1 score for AU decreased from 0.5801 to 0.5724. These findings highlight the effectiveness of the Transformer Encoder in capturing and enhancing the model’s ability to model temporal dependencies and contextual information.

\section{Conclusion}\label{sec:conclusion}

In summary, our study on human emotion recognition, presented at the 8th Workshop and Competition on Affective Behavior Analysis in-the-wild (ABAW), introduces a novel approach that combines fine-tuned CLIP with the aff-wild2 dataset. By incorporating Temporal Convolutional Network (TCN) and Transformer Encoder modules, our model significantly outperforms baseline performance. These results underscore the effectiveness of our methodology in advancing continuous emotion recognition and its potential to enhance human-computer interaction.

{\small
\bibliographystyle{ieee_fullname}
\bibliography{arxiv}
}

\end{document}